\documentclass[11pt, a4paper, twocolumn]{article}

\usepackage[left=1cm,right=1cm,top=2cm,bottom=2cm]{geometry}

\usepackage{abstract}

\usepackage{graphics}

\usepackage{amsmath,amssymb,amsfonts}
\usepackage{algorithmic}
\usepackage{graphicx}
\usepackage{textcomp}

\usepackage[utf8]{inputenc}
\usepackage{amsmath}
\usepackage{amsfonts}
\usepackage{amssymb}
\usepackage{graphicx}

\usepackage{caption}
\PassOptionsToPackage{hyphens}{url}\usepackage{hyperref} 
\usepackage{comment}
\usepackage{array}
\usepackage{float}
\usepackage{enumitem}

\def\lcls{L_{\mathrm{cls}}}
\def\lbox{L_{\mathrm{box}}}
\def\lmask{L_{\mathrm{mask}}}

\def\BibTeX{{\rm B\kern-.05em{\sc i\kern-.025em b}\kern-.08em
    T\kern-.1667em\lower.7ex\hbox{E}\kern-.125emX}}
\begin{document}

\title{\textbf{Deep Learning based Automated Forest Health Diagnosis from Aerial Images}}
\author{Chia-Yen Chiang$^1$,
Chloe Barnes$^2$, Plamen Angelov$^1$, and Richard Jiang$^1$}

\date{$^1$LIRA, Lancaster University, Lancaster, LA1 4WY, UK \\
$^2$2Excel geo, Hall Farm 2, Sywell Aerodrome, Sywell, NN6 0BN}




\twocolumn[
  \begin{@twocolumnfalse}
    \maketitle
    \begin{abstract}
      Global climate change has had a drastic impact on our environment. Previous study showed that pest disaster occured from global climate change may cause a tremendous number of trees died and they inevitably became a factor of forest fire. An important portent of the forest fire is the condition of forests. Aerial image-based forest analysis can give an early detection of dead trees and living trees. In this paper, we applied a synthetic method to enlarge imagery dataset and present a new framework for automated dead tree detection from aerial images using a re-trained Mask RCNN (Mask Region-based Convolutional Neural Network) approach, with a transfer learning scheme. We apply our framework to our aerial imagery datasets,and compare eight fine-tuned models. The mean average precision score (mAP) for the best of these models reaches 54\%. Following the automated detection, we are able to automatically produce and calculate number of dead tree masks to label the dead trees in an image, as an indicator of forest health that could be linked to the causal analysis of environmental changes and the predictive likelihood of forest fire.\\

      \newline
      \newline
    \end{abstract}
  \end{@twocolumnfalse}
]


\maketitle

\section{Introduction}
\label{sec:introduction}
More and more evidence indicates that global climate change has led to deforestation and extinction of species over the past decade. The incremental changes in temperature have not only caused extreme events such as floods, drought and gales, but inevitably affect landscapes and whole ecosystems. In such a situation, species may struggle to adapt to changes \cite{Ancienttreeguide}. Moreover, it is reported \cite{Germanyforests} that 110,000 hectares of forest have died in 2018, which is around the size as 150,000 football pitches or one nineteenth of Wales, due to extreme heat and storms. 

Certain species whose habitats were living woods have lost their habitats due to deforestation and dying trees. On the other hand, organism or creatures that take dead wood as habitats (for example: fungus, bacteria, predatory beetles, and parasitic
wasps) increase their population. Pests such as bark beetles and cardinal beetle can nest in dead tree trunks, and then may move to healthy trees and kill them \cite{Germanyforests}.

In addition, the dry, inflammable nature of dead wood makes fire disasters happen more frequently \cite{Californiafires}. In California, poor management of dead wood in forests has been identified as the major cause of wildfire. Both brought huge costs since the fire destroyed neighboring healthy trees and spread fast to villages nearby. Forest fires also trigger severe smog; for example, in 2015, wildfire smoke in Indonesia caused an estimated 100,000 premature deaths \cite{wildfiresmokeinIndonesia}.

Because the of impacts on ecosystems, inhabitants and households mentioned above,
technology to monitor forest areas has been developed rapidly in the past decades.
Technologies such as remote airborne sensor or satellite imagery made forest monitoring less costly, and allows monitoring without human involvement. By using these technologies, forest experts and disaster specialists not only can monitor forests
in real-time, but also adopt preventive measures for natural disasters.

On the other hand, pest disaster, one of the natural disasters caused by climate change,
has recently drawn experts' attention. The Canopy Health Monitoring (CanHeMon) \cite{canhemon} project aimed to eliminate the danger to forests from pine wood nematode (PWN), and so previous research has been done to detect dead trees within forests, to prevent them from becoming vectors for PWN. A MaxEnt-based iterative image analysis algorithm, applied to remote sensing data, as able to detect individual declining tree crowns with high accuracy \cite{canhemon}. The algorithm searches for the distribution with the maximum entropy for a given input set of occurrences and a set of constraining variables. As higher entropy indicates high disorder and abundance of information,
the algorithm can filter out irrelevant variables, leaving the most useful variables highly correlated to dead trees. However, the results revealed that the rate of false positives increases as the sample size decreases. Moreover, eliminating small size samples (smaller than 2.5m$^2$) harmed the true positive detection rate, dropping from 80\% to 65\%. This may indicate that the algorithm is not suitable to detect small objects or that there were not enough samples of small trees in the training data. 

In addition, the CanHeMon project suggested that using deep learning algorithms may provide a more generalised image analysis model and allow us to identify pixel-wise shapes of trees in an image. To investigate this, the CanHeMon team also performed tests using a neural network model. The results showed that rate of false positives was lower than for the MaxEnt-based algorithm, but the behaviour of the loss function indicated more epochs and iterations were needed in training the model. Hence, in this research, we extend the deep learning methods applied by the previous team.

In this study, we used a Mask Region with Convolutional Neural Networks (Mask-RCNN) algorithm, accompanied by transfer learning  \cite{Transferlearning} to deal with the problem of pixel-wise object identification and to enhance the learning pattern, i.e. through additional training sessions, the overall score performance can be improved by using pre-trained weights produced in the previous session. 

Moreover, to cope with small datasets, and problems with unlabelled data, we use synthetic tools: randomly assigning instances into backgrounds, both extracted from raw data. At the same time, we output COCO format annotations \cite{COCO} to be used in inference. In general, manually annotating these masks is quite time-consuming, frustrating and commercially expensive \cite{semanticsegmentation}. However, we were inspired by Adam Kelly, the founder of Immersive Limit, and were able to automatically create thousands of synthetic images and their annotations with minimal manual effort. \cite{CreatingCOCODatasets} \newline
The study makes four major contributions:

(1) Use synthetic data to do Mask RCNN modeling and reach about 54\% on mean average precision score (mAP) . The results show that it is possible to use synthetic methods to mitigate data scarcity.  

(2) Mask prediction is limited by selected ROIs and can be found from high loss of RPN. Therefore, applying semantic segmentation may be a more practical way than instance segmentation to detect dead tree region.\autoref{subsec:lossofRPN}

(3) Using two optimizers in training process caused loss converged faster than using one optimizer. \autoref{subsec:lossofRPN}

\begin{figure}
\centering
\includegraphics[scale=0.48]{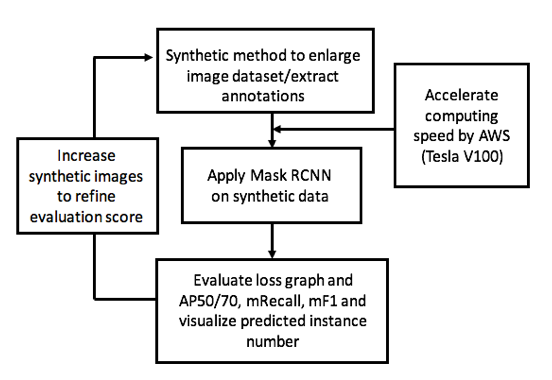}
\caption{Structure of the Mask-RCNN model.}
\label{fig:research}
\end{figure}

In addition,   \autoref{fig:research} provides our framework for forest analysis. More details will be in the following sections.

\section{Related work}

\subsection{Satellite imaging for forest analysis}
Although we did not use satellite images in our research, they have been used to monitor forest cover for a long time.
Analysing the change of forest cover generally relies on visual
interpretation of a dataset.
Accompanied with groud measurements, scientists are able
to calculate the loss of forest cover over time and
doing assess properties from the satellite imagery,
such as biophysical and biochemical properties of forest. \cite{groundmeasurements}

In addition, in various studies\cite{Landsatstudy1}\cite{Landsatstudy2}\cite{Landsatstudy3},  Landsat imagery\cite{LandsatDataAccess},
a popular open database of satellite imagery,
was sampled or trained to monitor forest cover changes.
Landsat imagery resolution is between 15-60m with multiple bands. 
"Landsat is a joint effort of the U.S. Geological Survey (USGS) and the National Aeronautics and Space Administration (NASA)...The data are useful to a number of applications including forestry, agriculture, geology, regional planning, and education. "\cite{Landsatgov}

\subsection{AI techniques in object detection}
With the remarkable advancement in artificial intelligence techniques in the past decades, there are several object detection algorithms which are now generally used, for example R-CNN\cite{RCNN}, Fast-RCNN\cite{Fast}, Faster-RCNN\cite{Loss}, YOLO\cite{yolo}, SSD\cite{SSD}, and NAS-Net\cite{NASnet}. When performing object detection, we are interested in finding two things: where our target objects are located, and which categories they belong to\cite{Review}. Our goal is to draw bounding boxes around target objects in an image, and output a percentage indicating how certain we are that they belong to certain classes. Moreover, state of the art techniques allow real-time prediction of object localisation and classification, for example You Only Look Once (YOLO) and Single-Shot Detector (SSD). 

\begin{figure}
\centering
\includegraphics[scale=0.5]{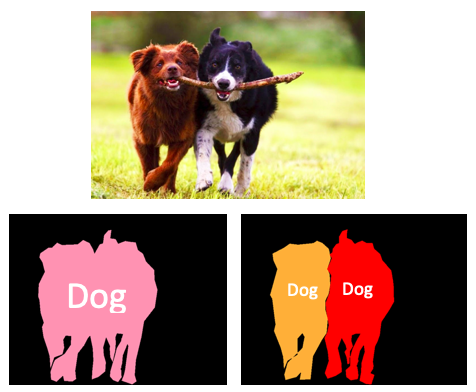}
\caption{Semantic segmentation (left) versus instance segmentation (right). Image from \cite{MaskRCNN}.}
\label{fig:segmentation}
\end{figure}

However, the cost of the fast inference time of such methods may be lower accuracy and low-resolution feature maps, due to a single network evaluation \cite{accuracycomparison}.. They use a single stage to complete both object localisation and classification, but in constrast, region based algorithms (RCNN, Fast-RCNN, Faster-RCNN and Mask-RCNN) separate the process into to two stages: first using a region proposal network (RPN) to propose a bounding box for a candidate object; then predicting the class and box offset, and outputting a binary mask for each region of interest’s ROI \cite{MaskRCNN}. That means we calculate both the classification scores and bounding box regression in second stage to decide which ROIs have the highest probability to contain target objects, and simultaneously perform instance segmentation on selected ROIs. Here, instance segmentation \cite{MaskRCNN} means to differentiate each target object and cut them apart even when they are labeled in the same class. It is often compared to semantic segmentation, labeling on each pixel with the class of its enclosing object or region (see \autoref{fig:segmentation}). Note that we only took segmentation into account because it was more precise to cut off dead tree regions in pixel-wise manner than simply output bounding boxes; for example, some boxes may contain a big proportion of living trees. 

\subsection{CNN and FCN in pixel-wise dead forest mapping}

Recent studies \cite{Mappingdeadforest1}\cite{Mappingdeadforest2}\cite{Mappingdeadforest3}\cite{Mappingdeadforest4}showed CNN and FCN were used in sementic segmentation on forest mapping.  In \cite{Mappingdeadforest1} a VGG neural network (an architecture of CNN) was proposed to classify pixels with a sliding window technique. One notable approach called ensemble learning was introduced to aggregate different predictions generated by the same model. This produced determistic estimates and reduced uncertainties caused by weight initialization settings. The resulting global accuracy predicting health status of two tree species was around 80\% to 85\%. However, dead trees were misclassified (defined as "omission error") at a rate of 35\% to 40\%.  

Another study suggested sliding windows \cite{disadvantagesofslidingwindows} cause high computational cost because neighbour pixel information, which contains the same information as centre pixels, is used to classify centre/target pixel. An algorithm such as FCN is developed to prevent duplicate computation by predicting many pixels at once. In addition, FCN can take various input sizes and output a dense prediction. 

A DenseNet (an architecture of FCN) proposed in \cite{Mappingdeadforest2} has outstanding results for classifying dead trees and standing dead trees. The resulting recalls and precisions are mainly within 90 to 100\% although a recall of a subset was 67.8\%. The author also encouraged that instance level prediction can be done to improve future research.   
       
\subsection{Transfer learning}

In the field of image processing, there have been many studies using transfer learning. The technique uses an existing trained network to improve the training accuracy for a new model. 

Since an outstanding trained network such as COCO or ImageNet has been trained using many images with a big number of labels, it seems sensible to reuse fundamental parts of the pre-trained net as low-level feature extractors, i.e. to recognize edges, colours and textures - whether detecting complex and subtle objects such as a human fact or simple objects such as rectangles, early layers in neural networks always detect the same basic shapes and edges. 

To be more specific, the COCO dataset\cite{COCO} is a large-scale object detection, segmentation, and captioning dataset. It has over 200\,000 labeled images with over 80 labels, while the ImageNet dataset has tens of millions of images with thousands of labels. \cite{Alexnet} 

\cite{Transferlearning} So to train a model on a custom dataset, all we need to do is replace the classification and bounding box predictors with customised ones which are able to predict our labels.
First, we take a pre-trained net without its FC part, since Fully Connected layers are used only for the old model's classification.
Next, a new FC containing customized label classifiers replaces the old.
Then, a convolution base which is initially close to the old classifier needs to be frozen since it contains high level features (for example cat ears, dog eyes, or a more representative feature) related to classification in the old model which may not be useful for our own customized classification problem.

\subsection{Transfer learning in forest aerial imagery}

We used a COCO or Imagenet pre-trained net and retrained the "head" parts for our model in the first session, i.e. we kept the COCO or Imagenet backbone but retrained the RPN, classifier and mask heads of the network. \cite{stepbystepprediction}

In the first training session, we only trained the head since we want to leverage the advantage from COCO and ImageNet, that is, use the weights that have been trained from many images. However, we found that merely training the head was not sufficient to get higher accuracy. Because of that, in the second and third sessions, we train all layers twice. Therefore, we denote the deeper model as “head + all + all”. 

Then in the second training session, we used the last weight produced in the first session to continue training. This gave us faster training times, and allowed us to use a well-trained model even when we had scarce data. 
Related studies on detecting objects in satellite imagery of urban landscapes\cite{CNNsatellite} show that by using an ImageNet pre-trained net, accuracy was increased by around 5\% for a fixed length of training time.
However, the study also showed a less satisfying performance even after using transfer learning because the model is trained to predict objects from another dataset which has different domain features, which may not be similar to our own training data.
Therefore, problems such as how transferable features in deep neural networks are, and how to effectively reuse pre-trained nets when most features of the target objects are different from those of the training data, have been discussed in \cite{DDT}.
A Selective Learning Algorithm (SLA) has been proposed in this case, which learns useful features from unlabelled data and introduces an intermediate domain to close the gap between target objects and training data.

\section{Preliminary on CNN and Mask RCNN}
\subsection{CNN}
\label{sec:CNN}

\subsubsection{What is a convolutional neural network?}
A convolutional neuron network (covnet, or CNN) is a class of deep neural network that has proved very powerful for image classification.  In 2012, Alex Krizhevsky used CNN to win the ImageNet competition which made neural nets come to prominence in the field of computer vision. \cite{Alexnet}
\subsubsection{Why use a convolutional operation?}

\begin{figure}
\centering
\includegraphics[scale=0.4]{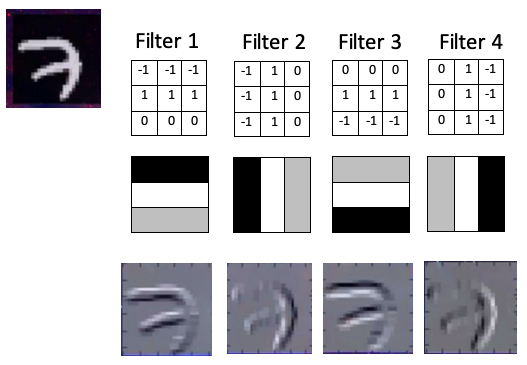}
\caption{Different feature maps produced by applying various filters to the same raw image. From \cite{featuremap}.}
\label{fig:featuremaps}
\end{figure}

\begin{figure}[h]
\centering
\includegraphics[scale=0.3]{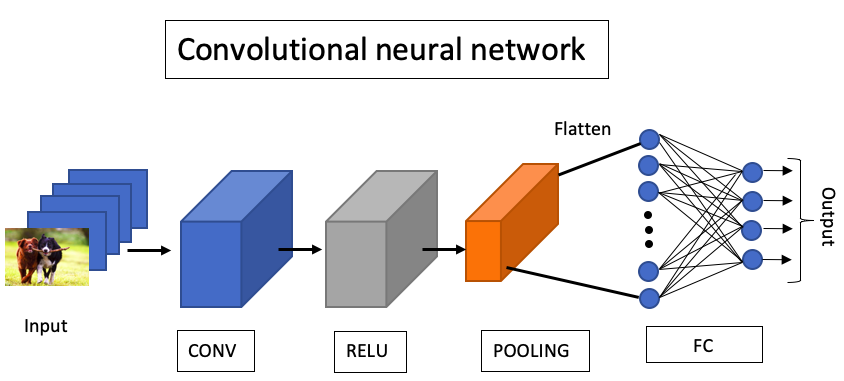}
\caption{Structure of a convolutional neural network. From \cite{ImageProcessing}.}
\label{fig:covnet}
\end{figure}

We can view an image as having 3D inputs with (height, weight, channel) so the input features have dimension H$\times$W$\times$C. For deeper layers, the channel value is usually increased in order to learn more features. A channel value is also the number of filters which contain special features we want to detect. In addition, given pre-defined filters, we can obtain corresponding feature maps by applying convolutional operation on input images. The operation not only shrinks the input dimension but preserve the spatial relationship between pixels; therefore a series of features are as well preserved. 
\autoref{fig:featuremaps} shows different feature maps produced by applying filters to the same raw images. 
Usually, a typical Convnet looks like \autoref{fig:covnet}. 

\subsubsection{Why can’t we use a simple CNN to predict object location?}

\begin{itemize}
\item The number of target objects and their coordinates in the raw picture is unknown before detection, so output nodes can’t be specified. In CNN, however, a fixed number of nodes is predefined for the output layer. Normally, outputs in CNN are fixed, predefined classes. Therefore, additional layers are needed to locate their coordinates.  

\item To classify if an object is a target or not, a number of fixed-sized input images are fed to CNN. However, in the object detection problem, input objects may have different sizes. In this case, a ROI pooling layer in RCNN is added to scale all objects in fixed size before passing them to the classification process.

\end{itemize}

\subsection{Mask RCNN structure}
\label{sec:maskrcnn}

\begin{figure}[h]
\centering
\includegraphics[scale=0.43]{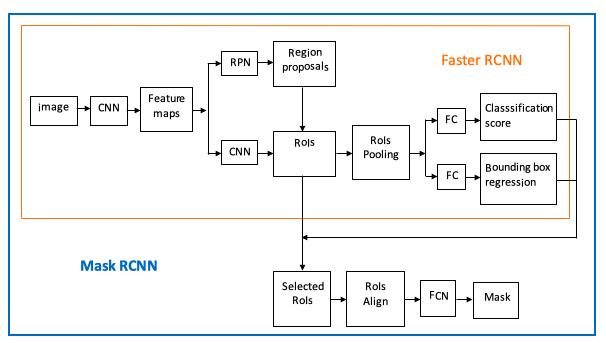}
\caption{Structure of a Mask RCNN model. From \cite{MaskRCNN}.}
\label{fig:fasterrcnn}
\end{figure}

In this section, we will first introduce Faster RCNN, the predecessor to Mask RCNN; then we will discuss Mask RCNN. This will give a broad picture of how Mask RCNN developed and the structure of the algorithm (see \autoref{fig:fasterrcnn}).

\subsubsection{Faster RCNN}

Faster RCNN begins with a standard convolution neural network (CNN) for feature extraction, which produces feature maps. A feature map is the output activation for a given filter in each layer. When a filter matches certain features on an image, the convolution operation \cite{convolutionarithmetic} returns high activation values and reveal those features in the feature maps. A feature map is also a mapping of where a certain kind of feature is found in the image \cite{featuremap}.

Then, it uses a Region Proposal Network (RPN) to extract Regions of Interest (ROIs) and resize the ROIs to a fixed size using ROI pooling. (Note: In section \autoref{sec: backbone generates feature maps} , we will talk about how RPN generates ROIs.) 

\begin{figure}[h]
\centering
\includegraphics[scale=0.6]{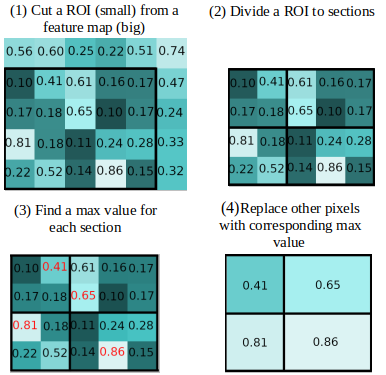}
\caption{How a max pooling layer works}
\label{fig:pooling}
\end{figure}

A pooling layer then reduces the dimension of the ROIs and the number of parameters in order to save training time and combat overfitting \cite{ImageProcessing}. It divides an ROI into a number of sections by pre-defined output size (Note: it is not necessary that ROI is perfectly divisible by the section number.) Next, a maximum value is found for each section and we replace all other pixel values in the section with this value. \cite{Loss} See \autoref{fig:pooling}.

Finally, the ROIs after max-pooling are flatten and fed to two fully connected layers (FCs). The flatten process makes pooled ROIs turn into vectors which store non-spatial features for later calculate the multilinear relationship between input neurons (in each ROI) and classes, and between neurons and bounding box size.

One FC layer calculates a bounding box regressor to refine the bounding box size, i.e. make each box covers the regions over its target object. Coordinates information will be sent to bounding box loss caculation. In next iteration, to minimize lost, bounding box coordinates will be updated therefore produce more accurate bounding box; and the other FC layer predicts a vector of probabilites over output classes using a softmax function. The probabilies will be used in calculation of cross-entropy loss which we adjusts probability information to minimize the cost in every iteration.   

\subsubsection{Mask-RCNN}

\begin{figure}
\centering
\includegraphics[scale=0.68]{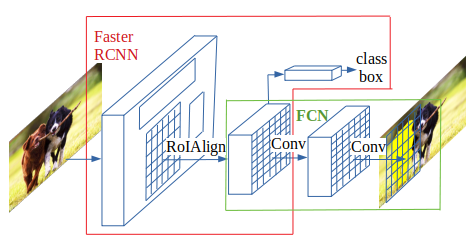}
\caption{Structure of the Mask-RCNN model. Image from \href{https://www.youtube.com/watch?v=FHytoCvj90w}{[link]}.}
\label{fig:maskrcnnstructure}
\end{figure}

Mask-RCNN can be thought of as a combination of a Faster RCNN that does object detection (class + bounding box) and a Fully Convolutional Network (FCN), a semantic segmentation method producing a pixel-wise mask. (See \autoref{fig:maskrcnnstructure})

In addition, the author of Mask RCNN \cite{MaskRCNN} did some fine-tuning of Faster RCNN and FCN. 
First, since the ROI pooling layer in Faster RCNN can lead to the location of target objects in the feature map being misaligned from their location in the original image, this is dealt with by refining the mask with an ROI Align layer.
ROI Align is a quantization-free layer, preserving exact spatial locations by replacing quantization with bilinear interpolation \cite{MaskRCNN}. Bilinear interpolation is used for computing the floating-point location values in the input \cite{DDT}.

\begin{figure}
\centering
\includegraphics[scale=0.6]{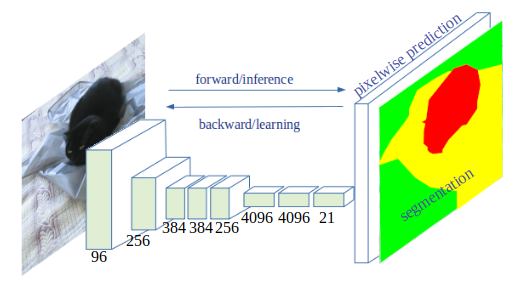}
\caption{FCN predicts a pixel-wise binary mask. Image from \href{https://ccsearch.creativecommons.org/photos/68cf9fb6-26a9-4a9a-a30c-923900617fed}{[link]}.}
\label{fig:fcn}
\end{figure}

Then FCN, a semantic segmentation method, is applied to each ROI in order to predict a pixel-wise, binary mask (see \autoref{fig:fcn}). Unlike the original setting of FCN, here the activation function in FCN was changed from softmax to sigmoid which was shown to significantly increase Average Precision (AP) \cite{MaskRCNN}. The reason is that FCN only calculates per-pixel sigmoid values and an average binary cross-entropy loss for a single, wholly classified object each time, rather than calculating per-pixel softmax and a multinomial loss for all pixels. Because of this, label prediction is decoupled with mask generation which avoids competition among classes.

\subsubsection{Backbone of Mask-RCNN}

Mask-RCNN uses Resnet-FPN as its default backbone and is a type of CNN which produces feature maps; the structure of Feature pyramid networks (FPN) allows us to build high-level semantic feature maps at different scales. When images are fed into FPN, first they go into the bottom-up process: the process functions the same as classic covnet extracts features, four feature maps (C2, C3, C4, C5) are produced and get reduced size (1/2 the size of the previous map) as the level goes up. At the same time, more features are extracted on the upper level but have lower resolution because of the convolution operation process.

In addition to the loss of resolution introduced by this, high resolution feature maps from lower level are not used for detection because of the lack of sufficient information (for example, it is possible that only some edges are detected). These two disadvantages make detection of small objects very difficult.

\begin{figure}[h]
\centering
\includegraphics[scale=0.6]{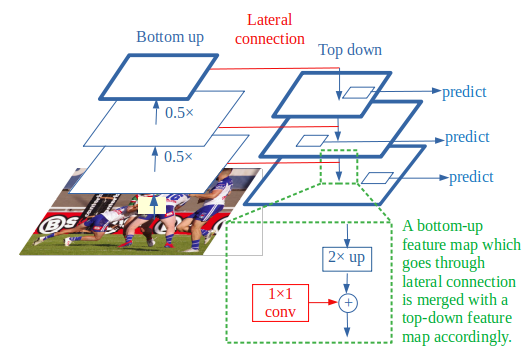}
\caption{Structure of Resnet-FPN, with the bottom-up pathway on the left, the top-down pathway on the right. \cite{FPN}. Image from \href{https://search.creativecommons.org/photos/95dc30f2-9394-4c28-96ab-ad2dc0bf57c0}{[link]}}
\label{fig:resnetfpn}
\end{figure}

In the FPN paper \cite{FPN}, however, the author added a top-down pathway (see \autoref{fig:resnetfpn}) to reconstruct a high resolution from a semantic rich layer by an upsampling method. The method doubles the map size (twice the size of the previous map) as the level goes down, with the resulting feature maps denoted as (P2, P3, P4, P5). A bottom-up feature map which goes through a lateral connection is then merged with a top-down feature map accordingly. The connection simply applies a 1$\times$1 convolution filter on each bottom-up feature map; therefore, it retains map sizes but reduces channel depths. In addition, to deal with spatial aliasing (the resulting location of objects may be different from their locations in the raw image because of repeatedly downsampling and upsampling) in the merged maps,  a 3$\times$3 covolution filter is used to generate feature maps with the highest resolution and accurate spatial information; therefore we can use them to do final predictions. 

\section{Our Framework for Forest analysis}

In this section, we explain how we applied Mask-RCNN to predict the location of dead trees in forests. The framework structure can be referred to \autoref{fig:research}. All experiments are conducted using Tensorflow-GPU docker on a Mac Air with eight NVIDIA Tesla V100 GPUs in Amazon Web Services (AWS). 

\subsection{Dataset collection and pre-processing}

The dataset was collected by aerial photography on May 15th 2019, from the Wood of Cree in Scotland, and was stored in a 10 GB merged tiff file.
As the file was too large for much of our software to process, we divided it into 40$\times$10 patches, each of 800$\times$800 pixels, using the Gridsplitter plugin for the QGIS software.
We then created our synthetic dataset from the raw data: we extracted over 300 images of dead trees as foreground objects and randomly resized, rotated and changed the lightness before they were placed into 63 candidate backgrounds, also extracted from the raw data. 
Next we produced a mask by filling random colours over the dead trees and black for the background. We then retrieved $x$ and $y$ coordinates of the target objects and stored them in a COCO format annotation file. 
Eventually, the dataset was extended from 225 patches to 5000 patches and produced 5000 annotations in only a few hours. 

\begin{figure}
\centering
\includegraphics[scale=0.28]{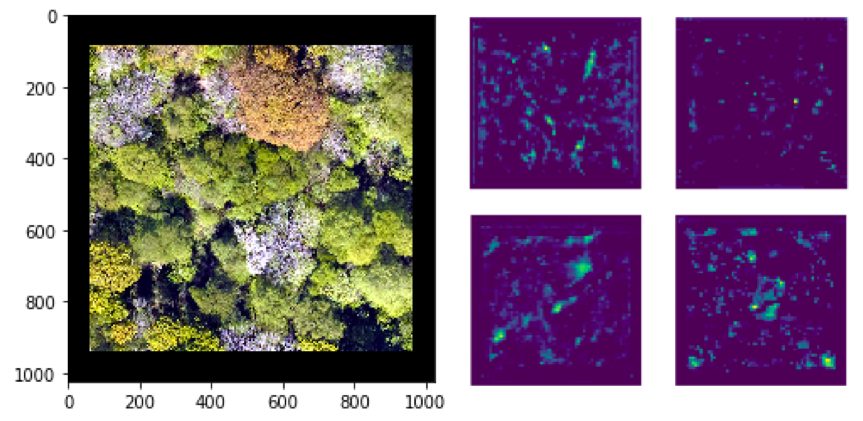}
\caption{A raw image, and the four corresponding feature maps. The lighter-coloured ``hot zones'' indicate the locations of the features we're interested in: shapes and colours of dead trees.}
\label{fig:forestfeaturemaps}
\end{figure}

\subsection{How the backbone generates feature maps}
\label{sec: backbone generates feature maps}

Once the dataset is fed to our convolutional backbone, it produces four feature maps for each picture in the dataset (see \autoref{fig:forestfeaturemaps}).

The feature maps are then scanned by RPN in a sliding-window fashion, finding anchors (pre-defined bounding boxes with fixed height and width) of high objectness score, i.e.\ the probability that an anchor contains an object or a non-object. Postive anchors have high objectness scores when they contain an object. 

Next, they are measured by value of 0.7 or more for the "Intersection over Union" (IoU) score. IoU is calculated as the ratio of the area of the overlap (intersection) between the pre-defined bounding box and the ground truthbox with the total area covered by both (union). 

\begin{figure} [h]
 \includegraphics[width=0.3\linewidth]{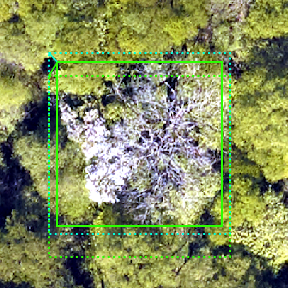}
 \hfill
 \includegraphics[width=0.3\linewidth]{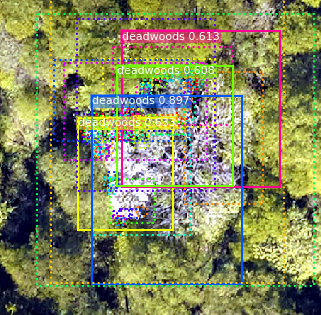}
 \hfill
\includegraphics[ width=0.3\linewidth]{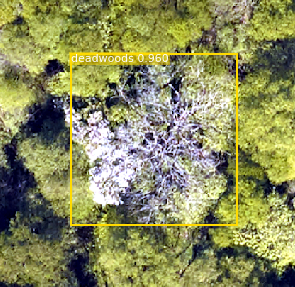}
\caption{Step by step detection from finding postive anchor boxes to trim down low confidence objects.}
\label{fig:anchorbox}
\end{figure}

Note that because positive anchors do not always include the whole boundary of their target object, they are examined by a regressor to refine the anchor location. However, if several positive anchors overlap too much, we keep the one with the highest objectness score and discard the rest. This method is referred to as Non-max Suppression.  \autoref{fig:anchorbox} shows the steps to merge the anchor boxes.

Before the region proposals are sent to the second stage, the algorithm scans and cuts out regions of interest (ROIs) from feature maps, as suggested by region proposals. ROIs are then scaled to a pre-defined, reduced dimension through the ROI pooling layer. (see \autoref{fig:pooling} ) 

\begin{figure} [h]
\centering
\includegraphics[trim=155 120 150 165, clip,width=0.45\linewidth]{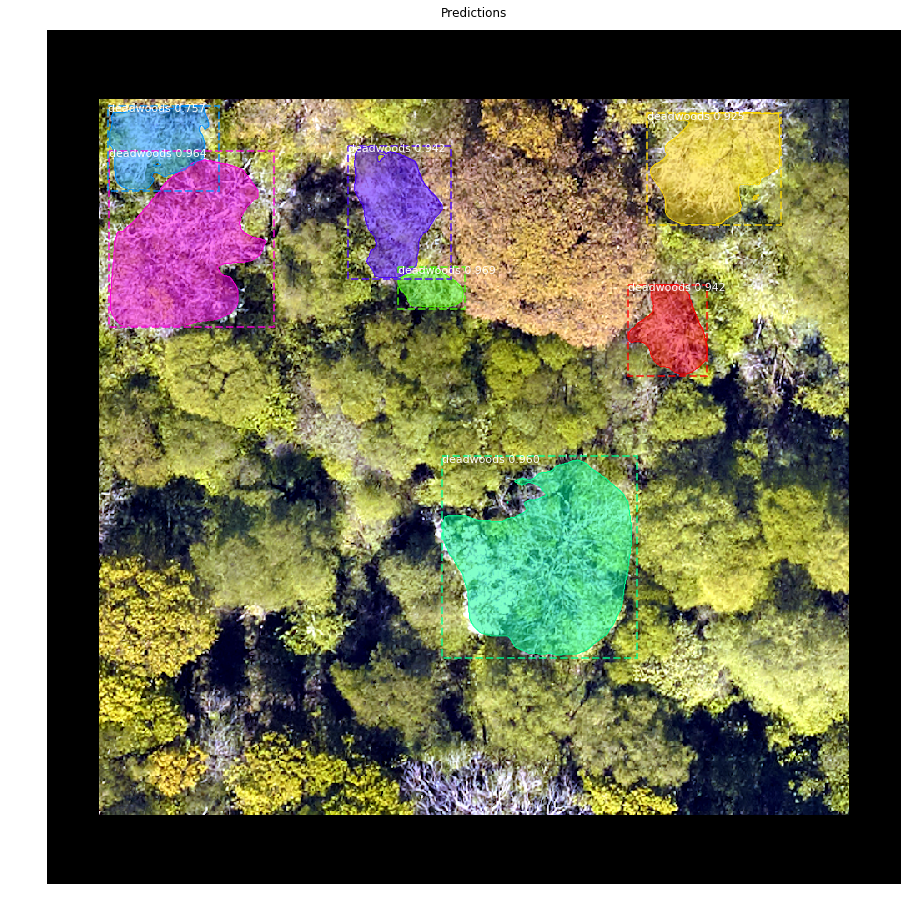}
\caption{Mask-RCNN prediction result.}
\label{fig:maskresult}
\end{figure}

Simultaneously, the selected ROIs are moved to the spatial locations corresponding to the original picture in the ROI alignment layer, and then are individually sent for pixel-wise classification in the FCN layer, resulting in a prediction result as in \autoref{fig:maskresult}.

\subsection{Model fine-tuning}
To make our training sessions faster and to more precisely predict target objects, we made some modifications to the Mask-RCNN model. 

First, we adopted a deeper model, i.e.\ training a ‘head + all + all’ model instead of a single-head model, and we redefined the learning rate for each training session. We set different learning rates based on optimizers, as in \cite{learningrate}. Our choices of parameters are summarised in \autoref{tab:finetunesettings}. Column with bold type shows the a new, fine-tune value is assigned, compared to its last model. Models are seperated to 4 groups with more similarity within group and are allowed to be compared.  

\begin{table}[h]
\resizebox{\columnwidth}{!}{%
\begin{tabular}{|l||l|l|l|l|l|l|}
\hline
Model & \begin{tabular}[c]{@{}l@{}}Pre-\\ trained \\ net\end{tabular} & \begin{tabular}[c]{@{}l@{}}Acti-\\ vation\\ func-\\ tion\end{tabular} & \begin{tabular}[c]{@{}l@{}}Learn-\\ ing \\ rate\end{tabular} & \begin{tabular}[c]{@{}l@{}}Opti-\\ mizer\end{tabular} & \begin{tabular}[c]{@{}l@{}}Drop-\\ out\end{tabular} & \begin{tabular}[c]{@{}l@{}}Data-\\ set\end{tabular} \\ \hline\hline
\begin{tabular}[c]{@{}l@{}}Fine-\\ tune-1\end{tabular} & \begin{tabular}[c]{@{}l@{}}Image-\\ Net\end{tabular} & Relu & \begin{tabular}[c]{@{}l@{}}$10^{-4}$,\\$10^{-5}$,\\ $10^{-6}$\end{tabular} & Adam & Yes & 2300 \\ \hline
\begin{tabular}[c]{@{}l@{}}Fine-\\ tune-2\end{tabular} & \begin{tabular}[c]{@{}l@{}}Image-\\ Net\end{tabular} & \begin{tabular}[c]{@{}l@{}} \textbf{Leaky}\\ \textbf{ Relu}\end{tabular} & \begin{tabular}[c]{@{}l@{}}$10^{-4}$,\\$10^{-5}$,\\ $10^{-6}$\end{tabular} & Adam & Yes & 2300 \\ \hline\hline
\begin{tabular}[c]{@{}l@{}}Fine-\\ tune-3\end{tabular} & \begin{tabular}[c]{@{}l@{}}Image-\\ Net\end{tabular} & \begin{tabular}[c]{@{}l@{}}Leaky\\ Relu\end{tabular} & \begin{tabular}[c]{@{}l@{}}$10^{-4}$,\\$10^{-5}$,\\ $10^{-6}$\end{tabular} & Adam & Yes &  \textbf{3600} \\ \hline
\begin{tabular}[c]{@{}l@{}}Fine-\\ tune-4\end{tabular} & \begin{tabular}[c]{@{}l@{}}Image-\\ Net\end{tabular} & \textbf{Relu} & \begin{tabular}[c]{@{}l@{}}$10^{-4}$,\\$10^{-5}$,\\ $10^{-6}$\end{tabular} & Adam & Yes &  3600 \\ \hline\hline
\begin{tabular}[c]{@{}l@{}}Fine-\\ tune-5\end{tabular} &\textbf{COCO} & Relu &\begin{tabular}[c]{@{}l@{}}$10^{-4}$,\\$10^{-5}$,\\ $10^{-6}$\end{tabular} &\textbf{SDG} & Yes & 3600 \\ \hline
\begin{tabular}[c]{@{}l@{}}Fine-\\ tune-6\end{tabular} & COCO & Relu & \begin{tabular}[c]{@{}l@{}}$10^{-4}$,\\$10^{-5}$,\\ $10^{-6}$\end{tabular} & SDG &\textbf{No} & 3600 \\ \hline
\begin{tabular}[c]{@{}l@{}}Fine-\\ tune-7\end{tabular} & COCO & Relu & \begin{tabular}[c]{@{}l@{}}$\mathbf{10^{-3}}$,\\$\mathbf{10^{-3}}$,\\ $\mathbf{10^{-3}}$\end{tabular} & SDG & No & 3600 \\ \hline\hline
\begin{tabular}[c]{@{}l@{}}Fine-\\ tune-8\end{tabular} & \begin{tabular}[c]{@{}l@{}}\textbf{Image-}\\ \textbf{Net}\end{tabular} & Relu & \begin{tabular}[c]{@{}l@{}}$10^{-4}$,\\$10^{-5}$,\\ $10^{-6}$\end{tabular} & SDG & No & \textbf{4000} \\ \hline
\end{tabular}%
}
\caption{Comparison of settings for our fine-tuning of Mask-RCNN.Column with bold type shows the a new, fine-tune value is assigned, compared to its last model. Models are seperated to four groups with more similarity within group. }
\label{tab:finetunesettings}
\end{table}

Secondly, to avoid overfitting, we used data augmentation, i.e. the algorithm accepts the original dataset, randomly transforms it and returns only the new, transformed data \cite{Augmentation}\cite{rotate}. We applied rotation by 0, 90, 180, and 270 degrees, tuned brightness and contrast between 0.9 to 1.1, applied horizontal flipping and clipped corners. 

Thirdly, to make the loss function converge faster, we changed our activation function from Relu to Leaky Relu.
The latter avoids the problem with “dying neurons” which Relu activation functions suffer from: when a neuron “dies”, whichever input it receives, its output will be equal to 0; in that case a neuron loses prediction power and stops learning. Leaky Relu, however, has a small but non-zero gradient for inputs below 0.

Fourthly, we applied a dropout function in the FPN layer to randomly drop out input nodes before doing classification and predicting bounding boxes. This decreases random correlations within neurons. It serves a similar purpose as data augmentation, i.e. to prevent overfitting \cite{dropout}.

Fifthly, we changed the optimizer from stochastic gradient descent (SGD) to Adaptive Moment Estimation (Adam)\cite{Adam}. An optimizer aims to update parameters $\theta$ in the negative gradient direction to minimize the loss. Studies show that Adam has less generalization error compared to SGD as well as being able to tolerate a noisy or sparse gradient. The algorithm is denoted
\begin{align*}
m_t &= \beta_1 m_{t-1} + (1 - \beta_1) g_t \\
v_t &= \beta_2 v_{t-1} + (1 - \beta_2) g_t^2
\end{align*}
where $m_t$ and $v_t$ are the decaying averages of the past gradient and past square gradient respectively, and the default value of each of the two $\beta$s is close to 1. We then update the estimates of both to deal with the bias caused by the initial averages $v=0$ and $m=0$:
\begin{align*}
\hat m_t &= \frac
{m_t}
{1 - \beta^t_1} \\
\hat v_t &=  \frac
{v_t}
{1 - \beta^t_2}.
\end{align*}
Because we initialize averages with zeros, the estimators are biased towards zero. However, the author of Adam optimizer used the above formulas to correct the estimators. \cite{Adam}

Then, using the updated parameters, the Adam update rule is:
\[
\theta_{t+1} = \theta_t - \frac{\eta}{\sqrt{\hat v_t} + \epsilon}\hat m_t,
\]
where $\eta$ is the learning rate, and we choose (as part of our tuning of the model) $\epsilon = 10^{-8}$.

Finally, we compare prediction results from fine-tune experiments (ex, activation function, dropout function, optimizer) to model without fine tune by visualizing loss history against epoch and a table reveals mean average precison (mAP) \cite{mAP}. 

\subsection{Loss function of Mask-RCNN}

Loss functions are used to evaluate how well a model performs \cite{Loss} - in our case we aim to minimise a loss function by optimising weights for three different tasks: classification, bounding box prediction and mask prediction. We use a loss function which is the sum of a loss function for each of these tasks:
\[
L = \lcls + \lbox + \lmask
\]
where $L_{\text{cls}}$ is the same as in Faster RCNN:
\[
\lcls( P_i, P^*_i ) = - \frac{1}{N_{\mathrm{cls}}}\left[P^*_i \log( P_i ) + (1 - P^*_i) \log( 1 - P_i )\right]
\]
where $P_i$ is the \emph{predicted} probability that the $i$th anchor is a target object. In Matterport Mask RCNN, the total number of anchors is 256 (positive and negative). The $i$th anchor above ``$i$th'' means a random number from 1 to 256.

If an anchor contains target objects, the \emph{ground truth label} $P^*_i$ is equal to 1, and otherwise it is equal to 0. In addition, the function is normalised by  $N_{\mathrm{cls}}$, i.e.\ the mini-batch size. A batch is the number of training samples in a given iteration. Overall, the classification loss $\lcls$ is a log-loss over two classes (i.e.\ target object and not target object) which can be thought of as a binary classification loss (cross-entropy loss).
In addition, $\lbox$ is also as in Faster RCNN:
\[
\lbox = \frac{\lambda}{N_{\mathrm{cls}}} P_i \, R( (t_i - t^*_i )_{i \in \{ \mathrm{x,y,w,h} \})}
\]
where $R$ is a robust $L_1$ loss:
\[
R(\mathbf{t}^u - \mathbf{v}) = \sum_{i \in \{ x,y,w,h \}} \mathrm{smooth}_{L_1} ( t^u_i - v_i )
\]
and $t_i^{u}$ for $i \in \{ x,y,w,h \}$ are the position and dimension parameters of the predicted bounding box, $v_i$ the parameters of the ground truth box, $\lambda$ is a balancing parameter which is set to be around 10 in the paper (so that both losses $\lcls$ and $\lbox$ are roughly equally weighted), and
\[
\mathrm{smooth}_{L_1}( s ) = \left\{
\begin{matrix}
0.5 s^2 & \text{ if } |s| < 1, \\
|s| - 0.5 & \text{ otherwise.}
\end{matrix} \right.
\]

We also have parameterised coordinates $t_i$ and $t^*_i$, 
\begin{align*}
(t_i) &= ( t_x, t_y, t_w, t_h )\\
(t^*_i) &= (t^*_x, t^*_y, t^*_w, t^*_h)
\end{align*}
where each individual parameterised coordinate is calculated given the predicted bounding box, ground truth box and anchor box coordinates as
\[
\begin{matrix}
t_x = (x - x_\mathrm{a})/w_\mathrm{a}, & t_y = (y - y_\mathrm{a})/h_\mathrm{a},\\ \vspace{0.2cm}
t_w = \log( w/w_\mathrm{a} ),  & t_h = \log(h/h_\mathrm{a})\\ 
t_x^* = (x^* - x_\mathrm{a})/w_\mathrm{a}, & t_y^* = (y^* - y_\mathrm{a})/h_\mathrm{a}, \\
t_w^* = \log( w^*/w_\mathrm{a} ), & t_h^* = \log(h^*/h_\mathrm{a}),
\end{matrix}
\]
where $(x_\mathrm{a}, y_\mathrm{a}, w_\mathrm{a}, h_\mathrm{a})$ are the coordinates of the anchor box.

Finally, $\lmask$ is the average binary cross-entropy loss which is calculated when the $k$th mask is the associated ground truth mask:
\[
\lmask = - \frac{1}{m^2} \sum_{1 \leq i,j \leq m} \left[ y_{ij} \log y_{ij}^k + (1 - y_{ij}) \log (1 - y_{ij}^k) \right]
\]
where $m \times m$ is the mask dimension for each ROI in each class, $y_{ij}$ is the ground truth label (0 or 1) in cell $(i, j)$ , while $y^k_{ij}$ is the predicted probability that the same cell belongs to the $k$th class \cite{Loss}.

\section{Experiments}
\subsection{Implement on Cloud Server with GPUs}
A cloud server provides a virtual environment to perform computation which can be accessed remotely via internet. The major benefit is to overcome hardware limitation, for example, lack of disk space to store data or computing power to train models. In this study, in order to deal with the enormous amount of data and complicated algorithms, a cloud server(AWS) is chosen to provide the environment needed for the AI program (Mask RCNN). 

Once an AWS computing instance has been registered, a private key file is given access to cloud server. Then, a Tensorflow-GPU docker file is used to activate eight GPUs as well as giving the user a temporary environment with numerous pre-installed python libraries. After the training session is done, the post-trained files are transfered back to cloud server before closing docker environment and retrieve files to local computer. 

The training process takes about 4-6 hours for a dataset of 2000 to 4000 images.  

\subsection{Evaluation of prediction performance}

As well as the loss function, we use COCO mean average precision
(COCO mAP) to evaluate prediction performance \cite{mAP}.
This indicator is widely used in evaluating models for object detection.
Firstly, after IoU decides the ROIs, we calculate @mAP
based on the prediction performance of those ROIs
(ignoring objects with an IoU score lower than 50\%).
Then we calculate the mean precision, recall and F1.
Then the AP is calculated as the area under the precision-recall curve:
the so-called Area Under Curve (AUC).
The curve shows the trade-off relationship between precision and recall.
When the area is larger, the predicted results are more accurate
(numbers of false positives and false negatives are lower) and more
positive cases are detected (number of false nagative is lower).

In addition, in COCO AP, 
the AP averaged over IoU thresholds is calculated,
i.e. AP50, AP75 and AP95.
In COCO mAP, an average for a 101-point interpolated AP is calculated.
This means recalls are divided into 101 points and
their maximized precisions are averaged:
\begin{align*}
\mathrm{AP} &=
\frac{1}{101} \sum_{r \in \{ 0,0,\cdots, 1,0 \}} \mathrm{AP}_r \\
&=
\frac{1}{101} \sum_{r \in \{ 0,0,\cdots, 1,0 \}} P_{\mathrm{interp}}(r).
\end{align*}

To make an even more direct indicator,
we also calculate the ROI number of two types of target objects
(\autoref{fig:twotypes}) which will be revealed
in the experimental results.


\subsection{Experimental results and analysis}
\label{sec:results}

Because CNN can easily suffer from overfitting,
we started training our model with a dataset of 2\,000 images
and a validation set of 300 images,
to which we applied many transformations.
We then looked to increase the size of the dataset gradually
to find the training set which gives us the most general model.
We trained each model with 100 epochs with
different hyperparameter settings.

\begin{figure}[h]
\centering
\includegraphics[width=0.2\textwidth]{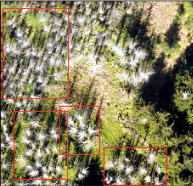}
\includegraphics[width=0.2\textwidth]{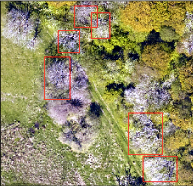}
\caption{The model ``Fine-tune-1'' can reveal the type A pattern (left) but not type B (right), and ``Fine-tune-2'' can reveal the type B pattern but not type A.}
\label{fig:twotypes}
\end{figure}

For the first pair of model, Fine-tune-1 and Fine-tune-2,
the results showed that for the two types of targets
(types A and B, see \autoref{fig:twotypes}) present in the data,
ROI detection did not perform well for predicting
type B objects (round shape) in the model ``Fine-tune-1''.
However, the results also showed the opposite trend in
model ``Fine-tune-2'', i.e. type B objects were detected reliably
but few type A objects were.
To resolve this, we decided to include more data in the next trial.

\begin{figure}[h]
\centering
\includegraphics[width=0.23\textwidth]{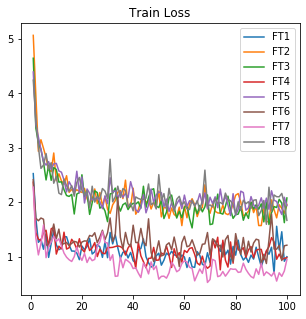}
\includegraphics[width=0.237\textwidth]{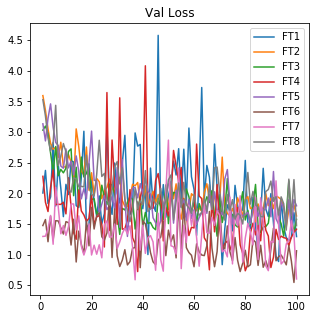}
\caption{Training and validation loss for all models.}
\label{fig:all-loss}
\end{figure}

\noindent For the next models,
we reveal the evaluation statistics in \autoref {tab:performances}.
Note that we make an abbreviation of "fine-tune" to "FT",
for example, ``FT3'' is the model ``Fine-tune-3''.
In addition, we indicate the best three performances
for each measurement in bold:

\begin{table}
\resizebox{\columnwidth}{!}{%
\begin{tabular}{|l||l|l|l|l|l|l|}
\hline
Model & FT3 & FT4 & FT5 & FT6 & FT7 & FT8 \\ \hline \hline
AP50 & 0.39 & \textbf{0.57} & 0.43 & \textbf{0.54} & \textbf{0.60} & 0.38 \\ \hline
AP75 & 0.14 & \textbf{0.53} & 0.13 & \textbf{0.48} & \textbf{0.56} & 0.10 \\ \hline
mPrecision & 0.62 &  \textbf{0.78}& 0.62 & \textbf{0.76}& \textbf{0.82} & 0.74\\ \hline
mRecall & 0.35 & \textbf{0.39} & 0.35 & \textbf{0.40} & \textbf{0.41} & 0.34 \\ \hline
mF1 & 0.29 & \textbf{0.32} & 0.28 & \textbf{0.33} & \textbf{0.34} & 0.28 \\ \hline
Type A (star) & 40 & \textbf{59} & 33 & \textbf{69} & 32 & \textbf{53} \\ \hline
Type B (round) & \textbf{71} & 24 & \textbf{64} & \textbf{74} & 24 & \textbf{64} \\ \hline
\end{tabular}%
}
\caption{Comparison of the performance of the models. The best three performances for each measurement are in bold. The final two rows show bounding box numbers for the two types of target objects.}
\label{tab:performances}
\end{table}
The table shows that results of FT6 are the best among all
of the fine-tuned models.
In addition, in \autoref{fig:all-loss} we plotted the loss results to see if FT6 had
the lowest training and validation loss.  

Because FT7 has the lowest training loss and
FT6 has lowest validation loss in \autoref{fig:all-loss},
we compared FT6 with FT7.

\begin{figure}
\centering
\includegraphics[width=0.236\textwidth]{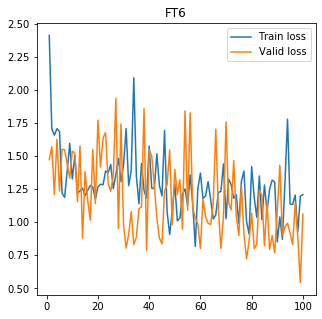}
\includegraphics[width=0.235\textwidth]{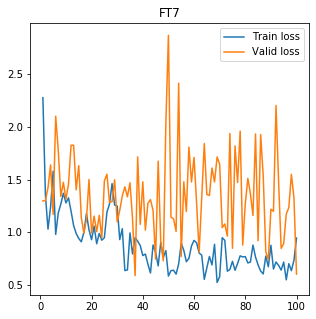}
\caption{Total loss for the models FT6 and FT7.}
\label{fig:total-loss-ft6-7}
\end{figure}

FT7 had a constant higher learning rate of $10^{-3}$
which is the closest of all our models to Matterport's Mask RCNN
default settings \cite{stepbystepprediction}, while FT6 followed
the fast-medium-slow learning rate of $10^{-4}$-$10^{-5}$-$10^{-6}$.
The results in \autoref{fig:total-loss-ft6-7} show that
FT7 suffered from both fluctuating loss and overfitting.
Overfitting happens when the training loss is less than the validation loss which means the algorithm learns local features appearing in the training set but not the validation set.
This indicates that a fast and fixed learning rate in FT7 has
no advantage over a learning rate which follows a convex pattern.

In addition, there were two major trends in \autoref{fig:all-loss}:
a big shift in the training loss, and two peaks in the validation loss.
Because all models had different settings,
especially in the different groups,
we assumed that the enlarged data set helped decrease the training loss.
In addition, for the validation loss,
two peaks were made by FT1 and FT4, and so we believe that this
may be because of their model setting (Relu + Adam + Dropout).

\begin{figure}[h]
\centering
\includegraphics[width=0.5\textwidth]{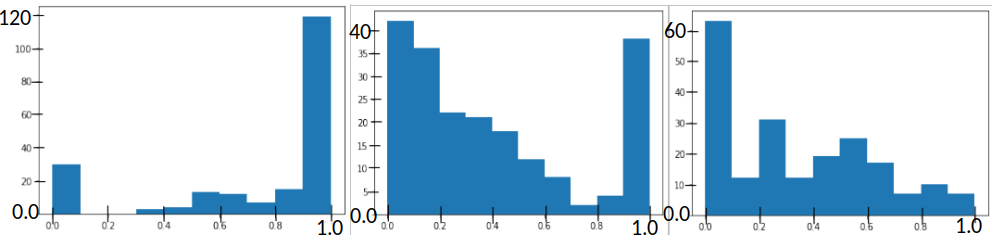}
\caption{
Distributions of precision, recall and F1 score for FT6 (respectively).
The x-axis counts the number of predicted bounding boxes while
the y-axis records the precision, recall and F1 scores respectively.
}
\label{fig:distribution}
\end{figure}

The distributions of precision, recall and F1
from the model FT6 are shown below in \autoref{fig:distribution}.
The diagrams show that FT6 has very high precision
but a rather low recall.
This means number of false negatives is higher than
the number of false postive.     

\section{Discussion}
\subsection{Advantages and disadvantages}
\label{sec: Advantage}
\label{sec: Disadvantage}

\textbf{Advantages of our method}
\begin{enumerate}[leftmargin=13pt]
\item Efficently produces images and annotations:

The synthetic method was designed by Adam Kelly to avoid
a time-consuming image collection and annotation process.
In Adam's experiment \cite{CreatingCOCODatasets},
he created 2000 images containing cigarette butts
and their annotations in 20 minutes from only 25 samples.
Because we had small numbers of data and no longer collected dead tree
images when we started to train our algorithm,
we chose the synthetic method rather than trying to find
more remote sensor data.

\item Less complex to apply compared to MaxEnt algorithm in CanHeMon project:

According to the image pre-processing section of the CanHeMon project,
there are 14 potential predictor variables related to texture layer,
containing significant image information.
Then, after a series of background and AUC setting,
the MaxEnt algorithm and principal component analysis are applied
to reduce the number of superfluous variables.
However, when applying CNN based algorithm,
the variables used as inputs and ouputs are rather simple:
raw images and their labels.
There is no manual elimination process for the variables/features
but the algorithm automatically adjusts feature weights after
backpropagation in each iteration.

\item False positive cases are rare:

The CanHeMon project suffered from a high rate of false positives,
which does not occur in our Mask RCNN-based model.
In the CanHeMon experiment,
researchers manually removed false postive cases and fed
the data back to the training session.
However, the mask results from our model rarely showed
false postive cases,
although there were still many areas of dead trees
which were not detected.

We found that a decrease in the IoU threshold can help
detect more dead trees, and therefore decrease the number
of false negatives.

Changing a hyperparameter in this way requires less manual effort
than manually dropping FP cases.
\end{enumerate}

\noindent\textbf{Disadvantages of our method}

\begin{enumerate}[leftmargin=13pt]
\item Data scarcity:

We gathered all of the images of dead trees we could from both online resources and our partner company 2Excel. However, we did not generate enough data to overcome the problems caused by data scarcity, as dead trees are an uncommon object, whereas most image datasets collect everyday objects (cars, people, etc.).
The main problem was that there were several shapes of dead tree in our raw data (see the two types in \autoref{fig:twotypes}) but we could not generate a sufficient number of images to deal with this variety.

\item The process of producing COCO format annotations did not give clear boundaries in synthetic images:
\label{item:synthetic-borders}

The synthetic method we used sometimes led to target objects being divided into more than one piece by another object overlapping them.
For example, see the red region in \autoref{fig:divided-region}.
We had to manually drop the label when this happened,
which is time consuming, reducing the benefit in time savings
we gained from using the synthetic method.
The overlapping type B
dead trees give us inaccurate bounding box predictions.

\begin{figure}
\centering
\includegraphics[width=0.2\textwidth]{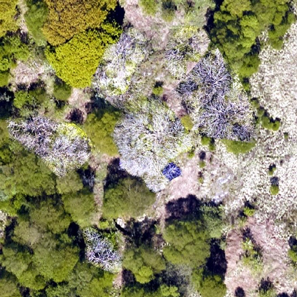}
\includegraphics[width=0.2\textwidth]{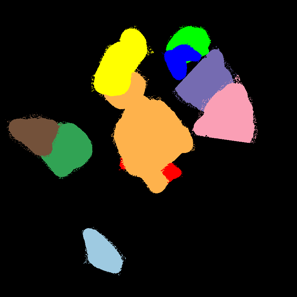}
\caption{A target object in this synthetic image is divided in two by another object placed on top of it.}
\label{fig:divided-region}
\end{figure}

\item Predicted bounding boxes failing to accurately cover a whole region of dead wood

Our synthetic method can make target objects overlap, and if they belong to type A (star shaped) then the algorithm will fail to distinguish the objects from each other, as in \autoref{fig:box-failure}. At best, they are bound to a larger group and given a single label.

\begin{figure}
\centering
\includegraphics{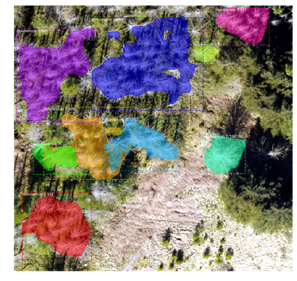}
\caption{Several target objects of type A being assigned to a single bounding box.}
\label{fig:box-failure}
\end{figure}

\item Reliability of mAP and other statistics

The mAP scores in our results were still not as reliable as the ones produced and evaluated using a raw (not synthetic) validation set. The synthetic validation set was necessary because of the scarce data, but it would be ideal to use only raw images for validation if possible.
\end{enumerate}

\subsection{Suggestions for future research}

In this research, we only trained our model with the company's aerial dataset instead of the Landsat dataset; however, we welcome to reseachers apply Mask RCNN to the Landset datset to make comparison with our results. In addition, we did not take into account other variables such as the forest phenology in our model, because the aerial dataset was collected in summer time (May 15th 2019) when living trees are easily distinguished from dead trees. However, we cannot exclude the possibility that Mask RCNN detects and labels as ``dead'' rare species of trees who naturally have white leaves. Because of that, we encourage researchers to apply other possible features in their models rather than only relying on an image processing algorithm.

Secondly, to mitigate the disadvantage of data scarcity, we enlarged our dataset and produced labels automatically using a synthetic method. However, we pointed out some issues with this method in \autoref{sec: Disadvantage}, although recent updates to the code we were using have dealt with some of the issues we saw in \autoref{item:synthetic-borders} of
the disadvantages in \autoref{sec: Disadvantage}.
As Adam Kelly suggested \cite{CreatingCOCODatasets}, it is still necessary to evaluate the advantages and disadvantages of using synthetic data to train a model.

The issue with bounding box inaccuracy is the major contributor to the high loss in RPN. \label{subsec:lossofRPN}
A possible solution is using a single FCN to classify each pixel if the instance-level segmentation is not necessary. Then, we could change the loss function from bbox + mask to FCN loss, decreasing the probability of high loss and enabling a more flexible
and accurate prediction because mask prediction is not limited by the selected ROIs' areas.

One other notable issue we experienced was how to increase the reliability of statistics 
(accuracy, mAP, mRecall and mF1) which were produced using the synthetic validations set to be consistent with the model's predictive power on raw images.  However, if we were able to collect a sufficiently large, manually annotated dataset, the statistics would be much more reliable. In addition, if the synthetic training set is replaced by manually-annotated raw data, it could increase the predictive power since the synthetic data may not include features which are present in the raw data.

\begin{figure}[h]
\centering
\includegraphics[width=0.236\textwidth]{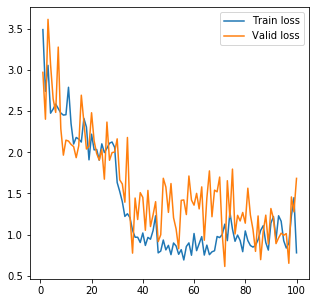}
\caption{The loss functions obtained by using an SGD optimiser until epoch 30, and then using ADAM.}
\label{fig:two-classifier-experiment}
\end{figure}

In the conventional training process, a single optimizer is used in a model. However, to take advantage of the merits of different optimizers, the author in \cite{ImprovingGeneralization01} used ADAM first, and when the loss stopped decreasing, replaced ADAM with SGD.
The results obtained by doing this were better than those obtained using either optimizer on its own. After reading this, we performed the same experiment with our model. As \autoref{fig:two-classifier-experiment} shows, replacing SGD with ADAM at the 30th epoch led to a sudden drop in the loss function. As it is hard to compare directly with the other models we tested, we did not include this model in \autoref{tab:finetunesettings}, but the results of our experiment are encouraging, so we would suggest that future research should look into using multiple optimizers for training.\\

\begin{figure}
\centering
\includegraphics[width=0.150\textwidth]{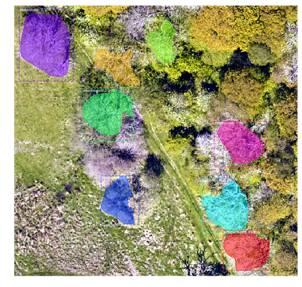}
\includegraphics[width=0.150\textwidth]{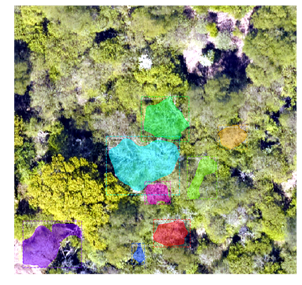}
\includegraphics[width=0.150\textwidth]{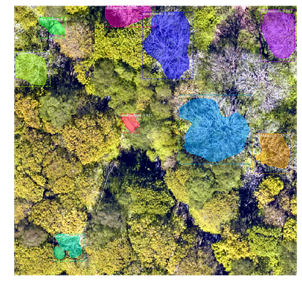}
\includegraphics[width=0.150\textwidth]{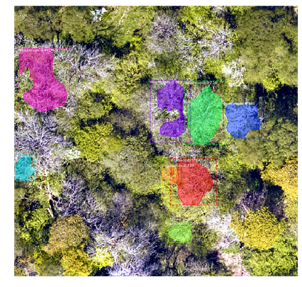}
\includegraphics[width=0.150\textwidth]{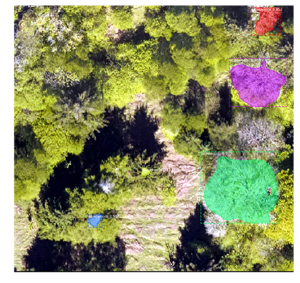}
\includegraphics[width=0.150\textwidth]{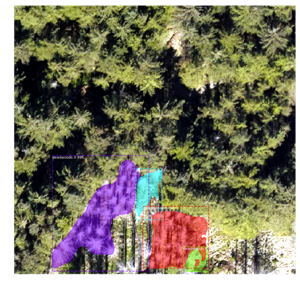}
\includegraphics[width=0.150\textwidth]{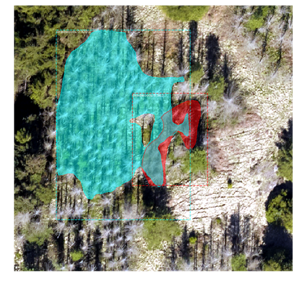}
\includegraphics[width=0.150\textwidth]{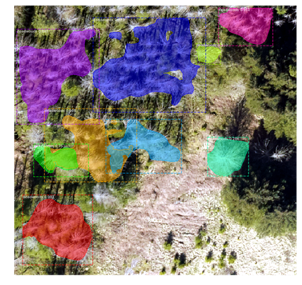}
\includegraphics[width=0.150\textwidth]{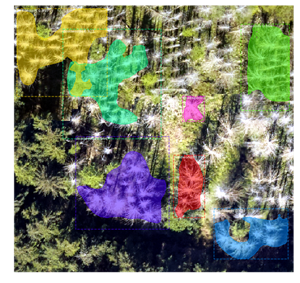}
\includegraphics[width=0.150\textwidth]{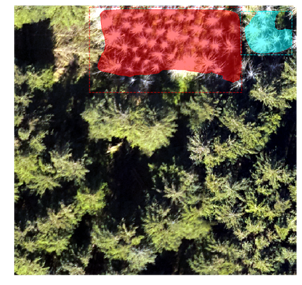}
\includegraphics[width=0.150\textwidth]{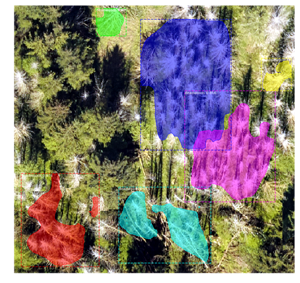}
\includegraphics[width=0.150\textwidth]{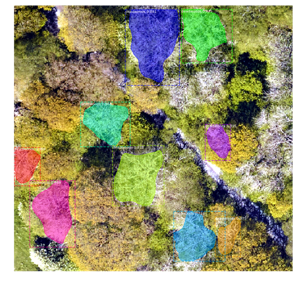}
\includegraphics[width=0.150\textwidth]{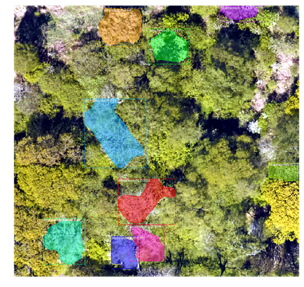}
\includegraphics[width=0.150\textwidth]{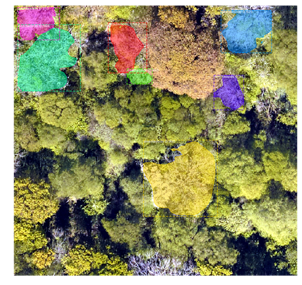}
\includegraphics[width=0.150\textwidth]{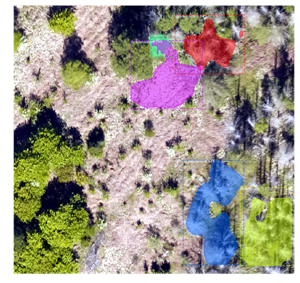}
\includegraphics[width=0.150\textwidth]{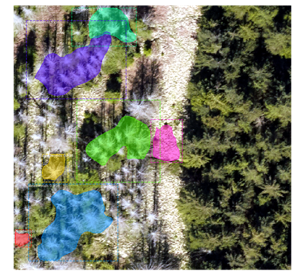}
\includegraphics[width=0.150\textwidth]{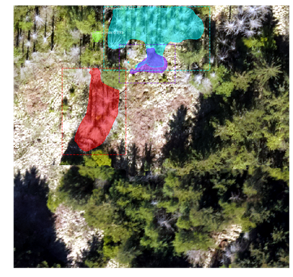}
\includegraphics[width=0.150\textwidth]{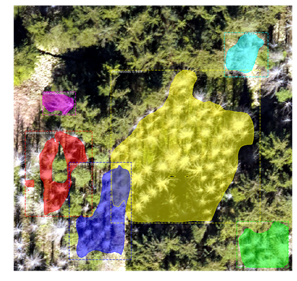}
\includegraphics[width=0.150\textwidth]{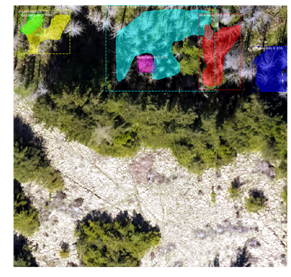}
\includegraphics[width=0.150\textwidth]{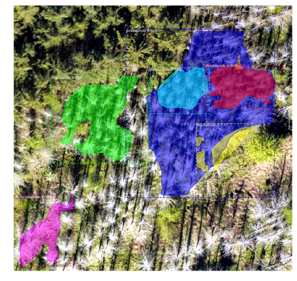}
\includegraphics[width=0.150\textwidth]{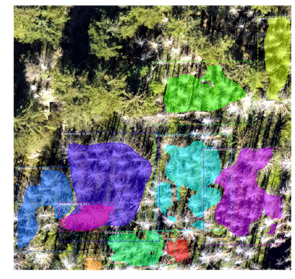}
\includegraphics[width=0.150\textwidth]{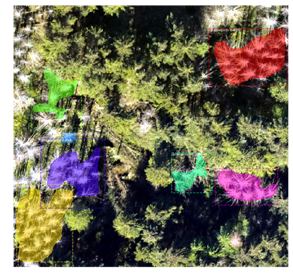}
\includegraphics[width=0.150\textwidth]{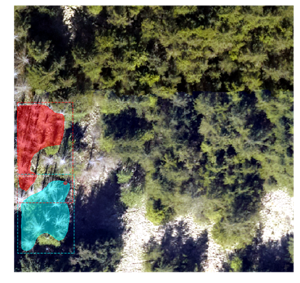}
\includegraphics[width=0.150\textwidth]{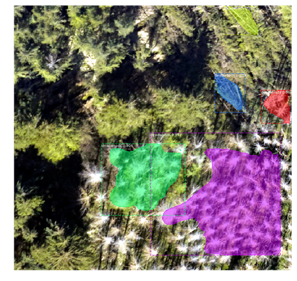}
\caption{More test results on aerial images.}
\label{fig:extra-examples}
\end{figure}

\autoref{fig:extra-examples} shows more of our test results. Mostly the detection framework works well on these images while the training process was carried out on the same dataset. However, there is yet a gap between the expected detection accuracy and the performance of our deep learning framework, and further improvement is needed to extend our framework to practical applications.

The limits can be ascribed to two aspects. First, deep neural networks has been reported not robust to noises and other similar perturbations (adversary attacks)\cite{jiang2019access01}, while DNNs are somehow lack of explicit explanability unlike typical statistic approaches that model the data space explicitly \cite{jiang2017pr}\cite{jiang2010tifs}\cite{jiang2010tbme}\cite{jiang2010tsmc}. Secondly, DNNs are date-driven and heavily rely on training datasets \cite{lecun2015nature}\cite{jiang2017htl}\cite{jiang2019access02}\cite{jiang2019access03}. In our experiments, we have a limited amount of data samples. A typical remedy to this issue can be based on the neural network models of Generative Adversarial Networks (GANs)\cite{GANs}, which can generate more similar images similar to real ones to improve the training process.

Targeting at dead wood detection, our work is based on the assumption that the dead tree dataset collected in the spring-summer period have different colors in leaves. However, this may vary according to tree species and some rare specious of trees may naturally have white leaves during four seasons. A future work can include forest phenology and classify the species of trees before the dead wood detection.

\section{Conclusion}
In conclusion, we present a new framework for automated dead tree detection from aerial images using a re-trained Mask RCNN approach, with a transfer learning scheme. We apply our framework to our aerial imagery datasets, and compare eight fine-tuned models. The mean average precision score (mAP) for the best of these models reaches 54\%. We also are able to automatically produce mask visualizations to label the dead trees in an image, so that the number of dead trees in a region can be automatically counted. Such aerial image-based forest analysis can give an early diagnosis of the conditions of forests and identify potential risk factors, and provide an potential solution to avoid of climate change caused disasters such as the recent forest fires in the California and Australia.

\section*{Acknowledgements}
This work was supported in part by the EPSRC grant (EP/P009727/1).



\bibliographystyle{IEEEtran}
\bibliography{IEEEabrv,library-of-references.bib}


\end{document}